\documentclass[10pt,twocolumn,letterpaper]{article}

\usepackage{ijcb}
\usepackage{times}
\usepackage{epsfig}
\usepackage{graphicx}
\usepackage{amsmath}
\usepackage{amssymb}
\usepackage{lineno}
\usepackage{subfigure}
\usepackage{epstopdf}
\usepackage{booktabs}
\usepackage{caption}
\usepackage{wrapfig}
\usepackage{makecell}
\usepackage{CJK}
\usepackage{txfonts}
\usepackage{bm}
\usepackage{color}
\usepackage{subfigure}
\usepackage{multicol}
\usepackage{multirow}
\usepackage{textcomp}
\usepackage[pagebackref=true,breaklinks=true,colorlinks,bookmarks=false]{hyperref}
\usepackage{hyperref}
\usepackage{indentfirst}
\ijcbfinalcopy 

\ifijcbfinal\fi

\makeatletter
\def\ps@IEEEtitlepagestyle{
	\def\@oddfoot{\mycopyrightnotice}
	\def\@evenfoot{}
}
\def\mycopyrightnotice{
	{\hfill \footnotesize copyright not found \copyright null\hfill}
}
\makeatother

\begin{document}

\title{Attention-guided Progressive Mapping for Profile Face Recognition}

\author{Junyang Huang\\
	South China University of Technology\\
	Guangzhou, China\\
{\tt\small eehjy1312@mail.scut.edu.cn}
\and
Changxing Ding \thanks{Corresponding author} \\
South China University of Technology\\
Guangzhou, China\\
{\tt\small chxding@scut.edu.cn}
}

\maketitle
\thispagestyle{empty}

\begin{abstract}
   The past few years have witnessed great progress in the domain of face recognition thanks to advances
   in deep learning. However, cross pose face recognition remains a significant challenge.
   It is difficult for many deep learning algorithms to narrow the performance gap caused by pose variations;
   the main reasons for this relate to the intra-class discrepancy between face
   images in different poses and the pose imbalances of training datasets. Learning pose-robust
   features by traversing to the feature space of frontal faces provides an
   effective and cheap way to alleviate this problem. In this paper, we present a method for progressively
   transforming profile face representations to the canonical pose with an attentive pair-wise loss.
   First, to reduce the difficulty of directly transforming the profile face features into a frontal one,
   we propose to learn the feature residual between the source pose and its nearby
   pose in a block-by-block fashion, and thus traversing to the feature space of a smaller pose by adding the learned
   residual. Second, we propose an attentive pair-wise loss to guide the feature transformation progressing in the
   most effective direction. Finally, our proposed progressive module and attentive pair-wise
   loss are light-weight and easy to implement, adding only about $7.5\%$ extra parameters. Evaluations on the CFP and
   CPLFW datasets demonstrate the superiority of our proposed method. Code is available at \href{https://github.com/hjy1312/AGPM}{https://github.com/hjy1312/AGPM}.
\end{abstract}

\let\thefootnote\relax\footnotetext{\mycopyrightnotice}

\section{Introduction}

\label{sec:introduction}
Progress in deep learning has significantly advanced the face recognition techniques,
such that the performance of many deep learning-based face recognition
algorithms can be comparable or even superior to human performance. In spite of this, a recent study has
demonstrated that the performance of many face recognition algorithms can drop by over 10$\%$ from frontal-frontal
to frontal-profile face verification \cite{author01}, while this gap is much smaller for human performance.
This indicates that large pose variation is still a challenge for the further advancement of face recognition,
particularly in unconstrained environments.

\begin{figure}
	\centering
	\includegraphics[width=0.95\linewidth]{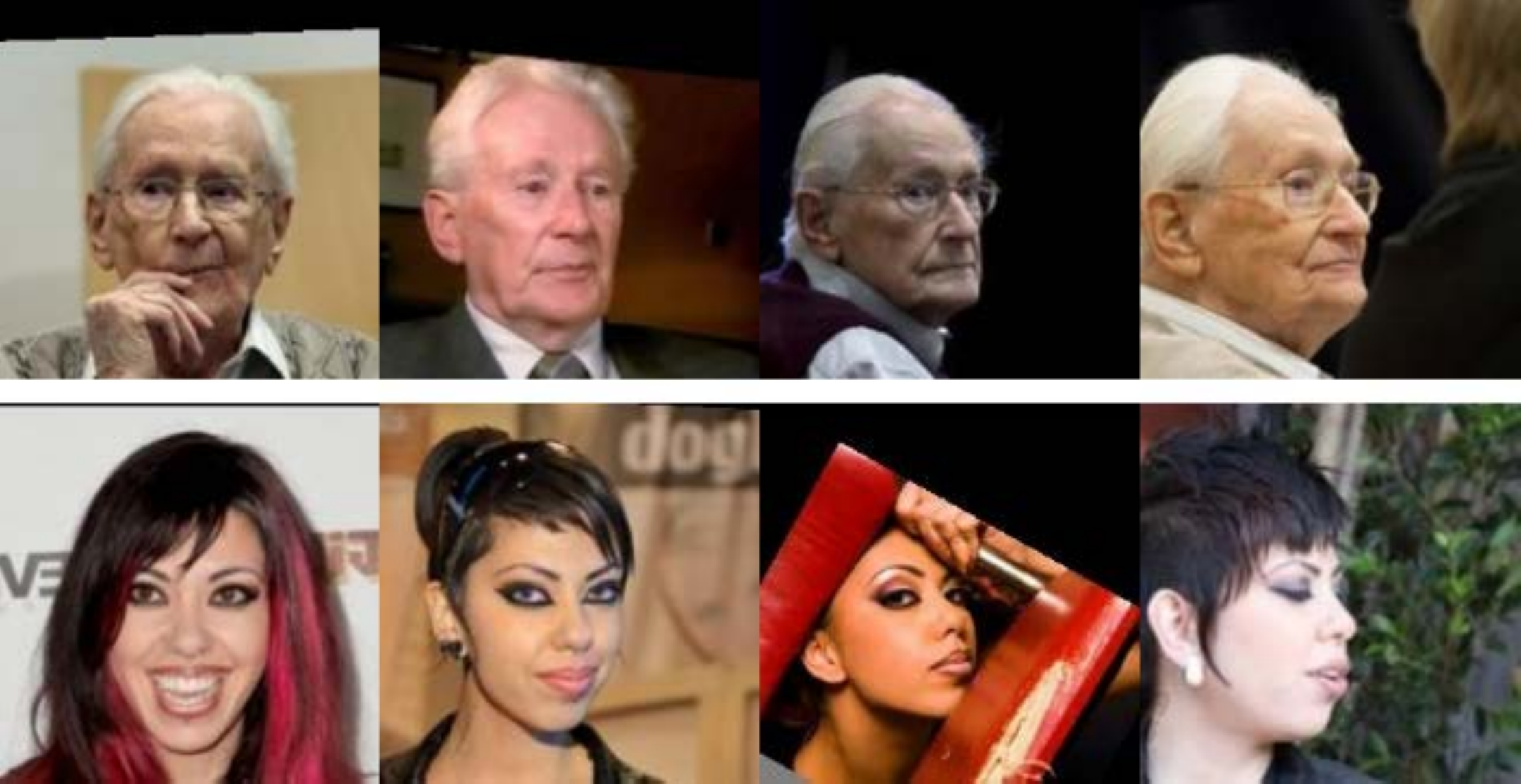}
	\caption{Sample face images in different poses. The images in each row are from the same person
		but with different head poses. It can be observed that obvious intrinsic facial information discrepancy
		arises when a face gradually turns from frontal view to profile view.
	}
	\label{vis_pos}
\end{figure}

There are two key reasons for the performance degradation caused by pose variation.
First, intrinsic information discrepancy exists between the frontal and profile face images.
As illustrated in Fig. \ref{vis_pos}, when the pose of face images change from frontal to profile, some parts
of faces are exposed, while others are occluded, thus appearance and texture changes inevitably
arise. This means that the information extracted from frontal faces and profile faces could be considerably
different. Second, deep learning algorithms for face recognition usually suffer from pose imbalance problems in massive datasets,
especially under real-world conditions. It can be quite expensive and infeasible to collect a large dataset
with a relatively even pose distribution. Since deep learning is heavily
data-driven \cite{author03,author82} and reliant on fitting the data, it is unsurprising that the algorithms
trained on these pose-unbalanced datasets fit and generalize
better on frontal face images than profile face images.

A large number of promising works have been developed to address this problem.
One primary research avenue is to synthesize face images across poses to reduce pose variation, including
normalizing face images to the canonical pose \cite{author07,author10,author53} and
augmenting training datasets by synthesizing face images across poses \cite{author47,author46}.
However, due to self-occlusion and the complex backgrounds, the synthesized face images may be not sufficiently photorealistic.
In contrast, some researchers have opted to tackle this problem in the feature space,
including learning a unified pose-robust feature mapping \cite{author57,author18,author20}
for face images at various poses and designing multiple pose specific models \cite{author48} to
formulate different feature mappings for various poses. However, due to the high nonlinearity
of pose variation, it is not easy to develop such a mapping that is capable of dealing with face images in all the poses.
Moreover, if handling these poses separately, extra computational cost would introduce.

Accordingly, in this paper, we propose a progressive pose normalization framework, together with an attentive pair-wise loss
to handle the pose variation problem by transforming features of the profile faces to frontal ones in a step by step manner.
As indicated in \cite{author18}, a gradual transition connection does exist between the profile
and frontal face domains in the feature space. However, directly modeling the transformation from profile to frontal face might
be problematic. As demonstrated in \cite{author26}, pose variations change smoothly but nonlinearly along a
latent manifold, directly transforming the features of extreme profile faces to those of frontal faces is a highly
nonlinear process, which requires searching for the optimal point in a large search region and might be trapped into local
minima. Meanwhile, multiple models are required to model these different nonlinear transformations at various poses.
Instead, if we decompose the task into several smaller progressive tasks with pose transformation limited
within a smaller interval, and ensure features in each interval possess similar properties, it would be much easier to achieve the goal.
Inspired by these observations, we propose to model the nonlinear and complex transformation from
extreme profile to frontal face by employing a progressive structure. More specifically, we roughly divide the pose into four
groups and design three stacking transformation blocks. Each block will convert the feature of larger
poses to that of its nearby poses, thus pose variation is narrowed down block by block.

We further introduce an attentive pair-wise loss (APL) to supervise this feature transformation
process. For many cross pose face recognition algorithms \cite{author18,author10,author46}, it is a natural choice
to employ frontal face images as a strong supervision signal. Motivated by these works,
we leverage the extracted features of frontal faces as the ground truth and penalize the L2 distance between them and the transformed features.
However, we note that it may be too rigid to enforce the transformed
feature to be completely the same as the frontal ones owing to the intractable content discrepancy
between profile and frontal faces. Instead, it would be better to focus only on maximizing their common elements in
the deep feature space. As it is quite difficult to identify these common elements, we propose the following easier approach.
Specifically, we apply features of frontal faces to generate an attention weights vector and then use them
to modulate the L2 loss as a form of channel attention, which
can capture the importance of each element in the feature vector. With the aid of these attention weights,
our network focuses on pushing profile features to be closer to frontal ones in the most effective channels
rather than all the channels, thus reducing the difficulty of frontalization and the
possibility of overfitting to trivial parts.

We conduct extensive experiments on two popular benchmark datasets for cross-pose face recognition, i.e., CFP~\cite{author01} and CPLFW~\cite{author29}.
The results show that our method consistently achieves superior performance for profile face recognition.


\begin{figure*}[tbp]
	\centering
	\setlength{\belowcaptionskip}{-0.4cm}
	\includegraphics[width=0.9\textwidth]{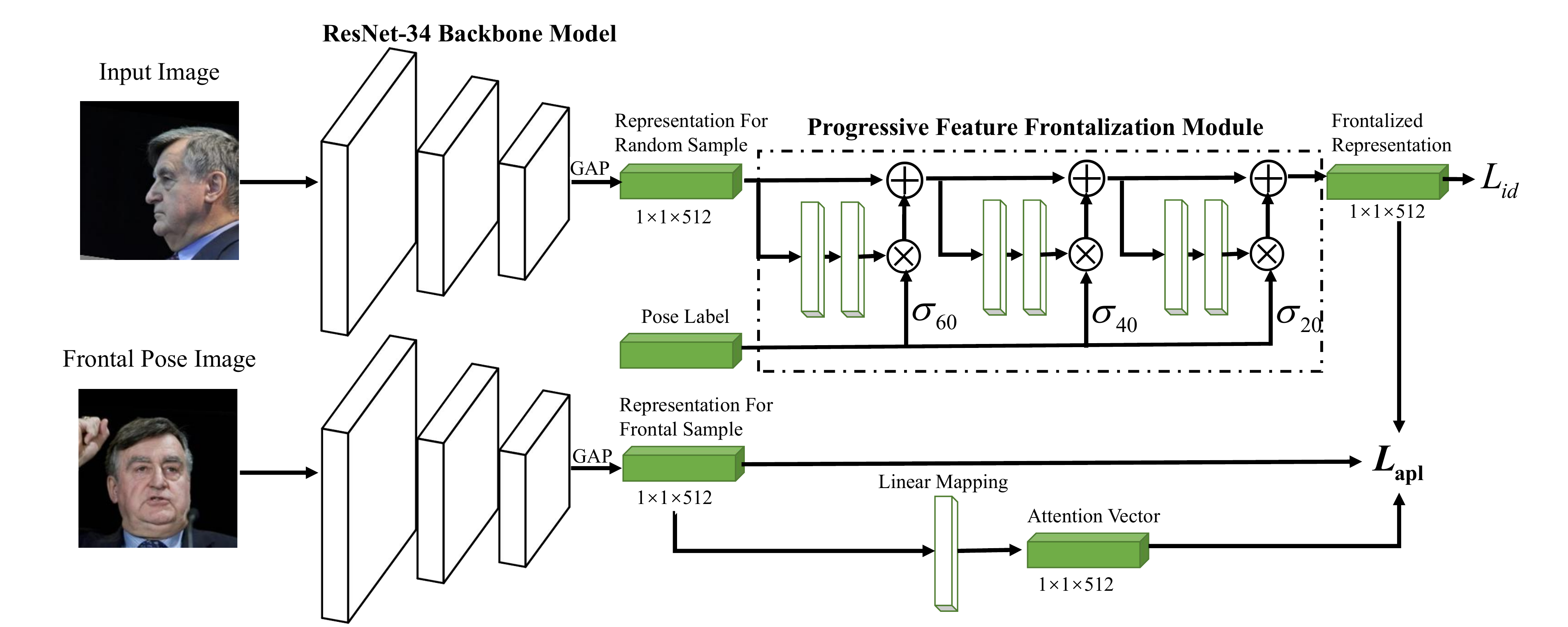}
	\caption{ An overview of our proposed approach. First, each training sample
		and its assigned frontal face image of the same identity are fed to the ResNet-34
		backbone to produce two feature embeddings. Second, progressive
		frontalization in the feature space are performed to map the feature embeddings
		into frontal feature space according to the pose of each image. Finally, with the modulation of
		an attention vector generated from the feature embedding of the assigned frontal face, identity
		classification loss and attentive pair-wise loss are utilized to train the entire network.
	}
	\label{overview}
\end{figure*}
\section{Related Works}
\label{sectionRelated Works}
\noindent \textbf{Pose Robust Face Recognition. }Enhancing algorithms' robustness to pose variation has long been a topic of interest in the field of face recognition.
One main line of inquiry is based on image generation. For example,
Qian et al. \cite{author10} introduced an unsupervised face normalization GAN model to transform a single face image of extreme
pose into a canonical view. Zhao et al. \cite{author47} proposed synthesizing profile face images
with a dual agent GAN framework for data augmentation. The drawback of these generative methods
is that it can be quite difficult to synthesize sufficiently photorealistic face images
due to the presence of occlusion and complex backgrounds. Another main research avenue is to learn a pose robust feature mapping.
For example, Iacopo Masi et al. \cite{author48} proposed a pose-aware framework that incorporates multiple models for various poses. Cao
et al. \cite{author18} introduced a Deep Residual Equivariant Mapping method to learn a residual between the features of profile faces and frontal
ones. In comparison, our proposed methods are much more light-weight and easier to implement,
increasing the total number of parameters by only $7.5\%$ across the whole network, while also being more effective for profile face recognition.

\noindent \textbf{Loss Functions for Face Recognition. } Loss functions
plays a key role in face recognition tasks. A large proportion of these functions are the variants of the classic cross entropy loss, including
SphereFace \cite{author32}, CosFace \cite{author33} and ArcFace \cite{author24}, et al. These loss functions
tend to minimize the intra-class distance and maximize inter-class distance by constraining the classification decision boundary.
Another mainstream approach is pair-wise loss. This approach aims at enhancing the compactness and discriminative power of extracted features by
intuitively forming pairs to train. For example, contrastive loss \cite{author37}
directly optimizes the Euclidean distance between similar and dissimilar pairs in Siamese networks.
Triplet loss \cite{author40} takes triplets as inputs and adds a margin between
positive pairs and negative pairs to reduce inter-class distance while increasing outer-class distance.
Nevertheless, in contrast to our proposed attentive pair-wise loss, which focuses on alleviating extreme
pose discrepancy to facilitate unconstrained face recognition, none of these loss functions explicitly aim at
handling the pose variation problem.

\begin{figure}[htb]
	\setlength{\belowcaptionskip}{-0.4cm}
	\centering
	\includegraphics[width=0.4\textwidth]{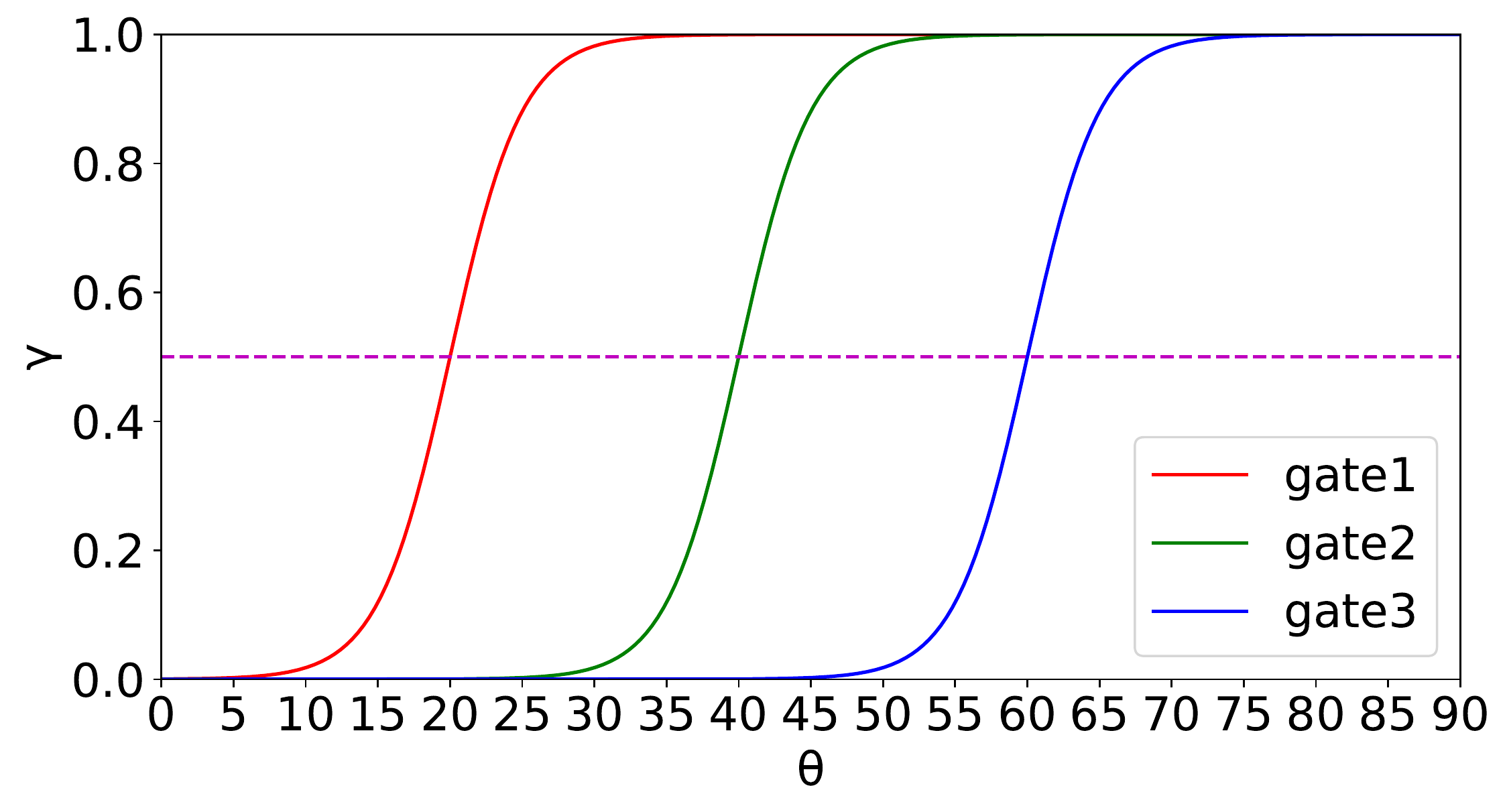}
	\caption{A visualization of the soft-gate mapping for the progressive blocks. The horizontal axis represents the value of
		the yaw angle $\theta$, while the vertical axis represents the soft gate coefficient $\gamma{\left(\theta_{i},\theta\right)}$.
	}
	\label{gate}
\end{figure}

\section{Method}
\label{Method}
In this paper, we propose a progressive feature frontalization framework together with an attentive pair-wise loss
for cross-pose face recognition. Fig. \ref{overview} presents an overview of our proposed method.

\subsection{Progressive Frontalization in Feature Space}
To begin with, we introduce the problem formulation. As illustrated in Fig. \ref{overview},
a face image $x_s$ and its accommodated frontal ground truth image $x_f$
are fed to the face recognition backbone to extract two deep feature embeddings,
$\psi\left(x_s\right)$ and $\psi\left(x_f\right)$, where the face recognition backbone
is denoted as a mapping function $\psi$. Our goal is to force $\psi\left(x_s\right)$ to
approach the feature space of frontal faces to the greatest extent through our proposed progressive module.

In the interests of clarity, we assume that face poses are divided into four bins within $\left[0^{\circ},90^{\circ}\right]$
according to the absolute values. Face images of $\left[0^{\circ},20^{\circ}\right]$, $\left[20^{\circ},40^{\circ}\right]$,
$\left[40^{\circ},60^{\circ}\right]$ and $\left[60^{\circ},90^{\circ}\right]$ are respectively denoted as frontal face, half
frontal face, half profile face and profile face, respectively.

In our method, each progressive block specifies a certain nonlinear feature transformation
between a pose bin and its nearby bin. For illustrative purposes, we term the three blocks respectively
as profile to half profile block, half profile to half frontal block, and half frontal to frontal block in
sequence. Each block takes the output of the last block as input, and then maps the features
at a larger pose to virtual features at its nearby pose, or simply performs an identity mapping when the
pose of the current input is smaller than the target pose of the block.

Similar to \cite{author18,author81,author80}, we utilize the residual block of our backbone as the basic architecture of the
transformation model; however, we opt to replace the $3\times3$ convolutions with
$1\times1$ convolutions and incorporate soft gates into these blocks, facilitating the adaptive
performing of suitable transformations for different poses. More specifically, each progressive block
attempts to learn a residual between features from
a larger pose to its nearby pose, which is then modulated by the pre-calculated soft gate coefficient and
plus the original features to formulate the features of the target pose. We denote these residual blocks respectively
as $R_i\left(i=1,2,3\right)$, while the soft gate mapping is denoted as $\gamma\left(\theta_i,\theta\right)$;
subsequently, the output of each progressive block is formulated as bellow:
\begin{equation}
\begin{split}
f_{i}\left(x\right)=f_{i-1}\left(x\right) + \gamma{\left(\theta_{i},\theta\right)}R_i{\left(f_{i-1}\left(x\right)\right)}. \label{pro_block}
\end{split}
\end{equation}

The coefficient $\gamma\left(\theta_i,\theta\right)$
serves as a selecting gate for the features that pass through each progressive block. This coefficient in the range of
$\left[0.0,1.0\right]$ determines the amount of the learning residual for the input feature of each block.
Given the features of a face image at the pose $\theta$ as input, each block compares $\theta$ with its preset threshold pose
value  $\theta_{i}\left(\theta_{i}=60^\circ,40^\circ,20^\circ\right)$, then produces a soft gate coefficient accordingly, as follows:
\begin{equation}
\begin{split}
\gamma{\left(\theta_{i},\theta\right)}=\frac{1}{1+e^{-10\left(\theta/\theta_i-1\right)}}. \label{yaw_corr}
\end{split}
\end{equation}

As shown in Fig. \ref{gate}, this formula can be also conceived as a soft binary mapping. When $\theta$ is slightly
larger than $\theta_i$, the soft gate coefficient will quickly saturate to 1.0; moreover, when $\theta$ is slightly
smaller than $\theta_i$, this coefficient quickly decrease to 0. It is noteworthy that as the pose variation
changes nonlinearly but smoothly, it would be unsuitable to directly employ a step function for mapping. Instead, a
transition zone is necessary when $\theta$ approximates $\theta_i$. Taking this into account, we employ
the soft gate mapping function presented above to control whether the blocks conduct frontalization transform or
directly perform an identity mapping.

\subsection{Attentive Pair-wise Loss}
To calculate the attentive pair-wise loss, we first assign a frontal face image of the same identity
for each input face image. Each input face image $x_s$ and its accommodated frontal face image $x_f$ are passed through
the same backbone model ResNet-34 and formulated feature embedding as $\psi\left(x_s\right)$ and $\psi\left(x_f\right)$ respectively.
As discussed above, it is difficult to determine the potential common elements in the embeddings of the image pair when pose difference is large,
as the extracted features of such profile face images tend to be less discriminative.
However, and notably, most common elements of the frontal-profile face image pair have strong responses
in the feature embedding of frontal faces. Thus, we formulate the attentive weighting vectors as follows:
\begin{equation}
A\left(x_f\right) = \frac{abs\left(\psi\left(x_f\right)\right)}{max{\left(abs\left(\psi\left(x_f\right)\right)\right)}}
,\label{gate_corr}
\end{equation}

where $abs\left(\cdot\right)$ represents the element-wise absolute value of a vector, and
$max\left(\cdot\right)$ returns the largest element in a vector.

With the frontalized feature embedding of profile face denoted as $F\left(\psi\left(x_s\right)\right)$,
the proposed attentive pair-wise loss can be calculated as follows:
\begin{equation}
\begin{split}
L_{apl}=\frac{1}{N}\sum_{i=1}^{N}{\lVert{\left(F\left(\psi\left(x^i_s\right)\right) - \psi\left(x^i_f\right)\right) \odot A\left(\psi\left(x_f\right)\right)}\rVert}^2_2
,\label{att_mse}
\end{split}
\end{equation}

where $N$ denotes the batch size, while $\odot$ represents the element-wise multiplication operation.

During training, we combine the cross entropy loss for identity classification and the above attentive pair-wise loss together to train
the entire network end to end. The overall loss function can be expressed as follows:
\begin{equation}
L=L_{id}+\lambda{L_{apl}},\label{total_loss}
\end{equation}

where $\lambda$ is a hyper parameter used to balance the cross entropy loss and the attentive
pair-wise loss.

\section{Experiments}
\label{sectionExperiments}
\subsection{Datasets and Evaluation Metrics}
\noindent\textbf{CFP} is a challenging in-the-wild dataset that focuses on face verification from frontal to extreme head
poses \cite{author01}. It consists of 500 celebrities, with 10 frontal face images and four profile face images for each person. In the
standard protocol, the whole dataset is split into 10 folds, each of which contains 350 intra-class pairs and 350
inter-class pairs, with the same number of frontal-frontal and frontal-profile pairs. Accuracy, Equal Error Rate (EER)
and Area Under Curve (AUC) metrics are then estimated by conducting 10-fold cross validation.

\noindent\textbf{CPLFW} is reorganized from the LFW dataset \cite{author73} by searching and selecting 3000 positive pairs with large pose difference
to emphasize intra-class pose variation, meanwhile also eliminating race and gender differences between 3000 negative pairs \cite{author29}.
The evaluation on CPLFW is based on the official protocol, i.e. 10-fold cross validation, with 300 positive pairs and 300 negative pairs for each fold.

\subsection{Implementation Details}
\noindent\textbf{Training Data}. We separately employ a subset of the MS-Celeb-1M \cite{author03} database and a refined clean version
of Ms-Celeb-1M proposed by \cite{author24} as our training data, in order to facilitate fair comparisons with other methods.
To create the first subset, we sampled 2,179,216 face images from 62,338 identities in the original MS-Celeb-1M dataset,
while the refined clean version contains 5,193,691 face images of 93,436 identities.
Moreover, in the interests of fair comparison, we have removed the images of identities that overlap with the testing datasets.

\noindent\textbf{Data Preprocessing}. We perform standard preprocessing on both the training datasets and testing datasets.
First, we detect the face bounding boxes and five facial landmarks with MTCNN \cite{author71} for each face image.
We then crop the faces with the detected boxes, align them with five facial
landmarks (the two eye centers, the nose tip and both mouth corners), and resize to 230 $\times$ 230 pixels.
Meanwhile, we adopt the tool in \cite{author72} to estimate the yaw angles of the face images.
Besides, common data augmentation methods are utilized to reduce over-fitting in the training stage, i.e. random gray scaling,
random horizontal flipping, and random cropping.

\begin{figure}[tbp]
	\setlength{\belowcaptionskip}{-0.4cm}
	\centering
	\includegraphics[width=3.3in]{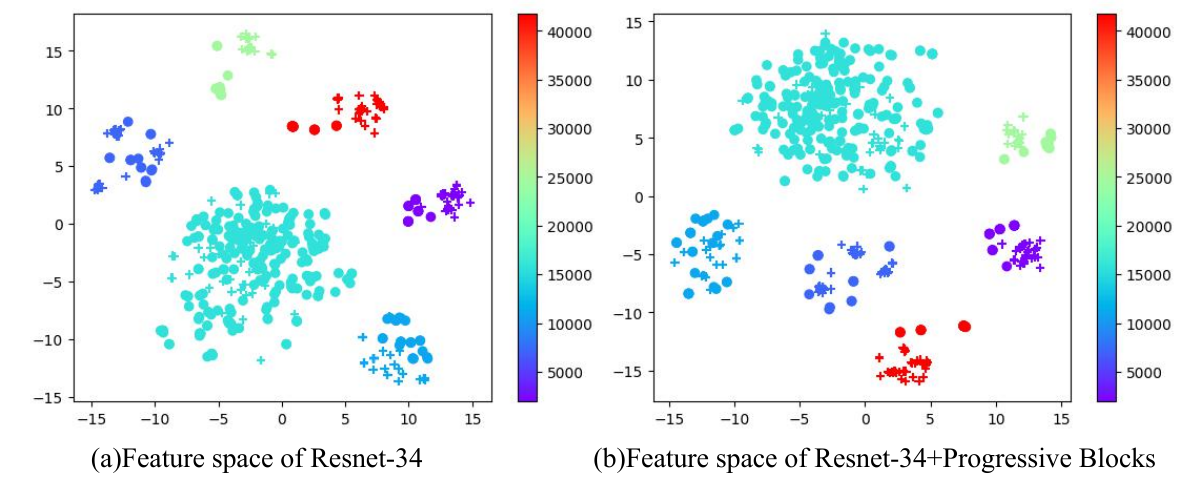}
	\caption{Visualization of feature cluster used TSNE(best view with zoom in). Here we use solid
		dots $\bullet$ to represent frontal faces and symbols + to denote profile
		faces. Different colors are used to represent different subjects. 
		(a) Feature space of the baseline, i.e. ResNet-34. (b) Embeddings
		obtained by ResNet-34 with the progressive blocks. 
		It is shown that the feature clusters become more 
		compact despite of pose variations. 
	}
	\label{Tsne}
\end{figure}

\begin{figure}
	\centering
	\includegraphics[width=0.95\linewidth]{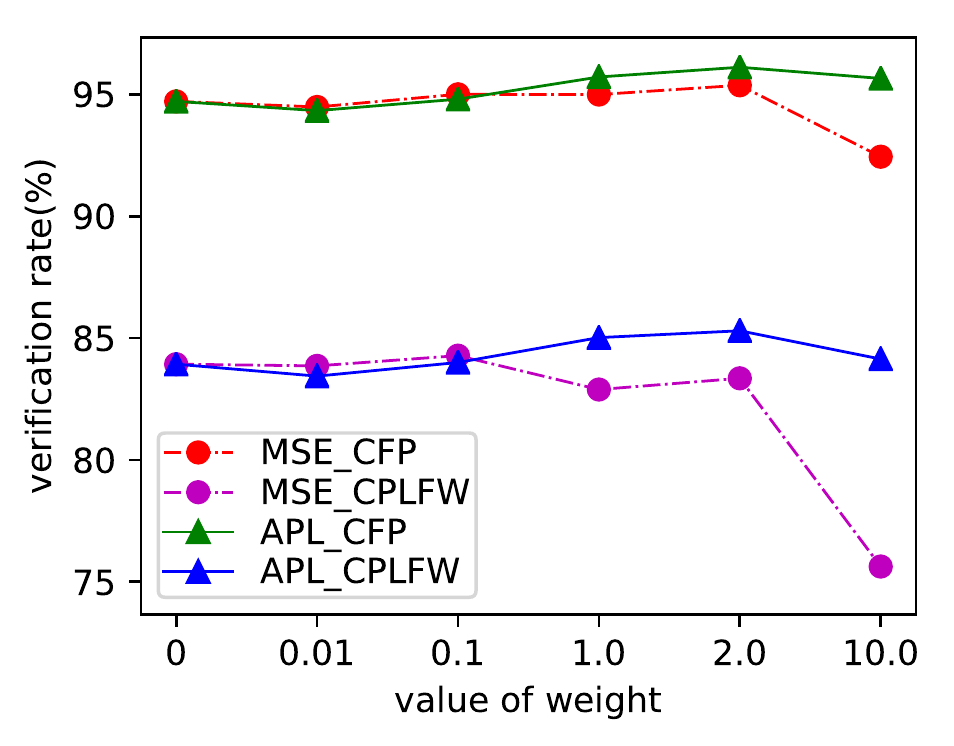}
	\caption{Comparison in verification accuracy between APL and plain MSE with different weights.
	}
	\label{mse_weight}
\end{figure}

\noindent\textbf{Training Setting}. Our framework is implemented in Pytorch.
We train our network on two NVIDIA TITAN V GPUs with a batch size of 200.
To construct a mini-batch for training, we first randomly sample face images from the
training set, and then respectively sample a frontal face image with a pose absolute
value smaller than $10^{\circ}$ as the ground truth for each of them. For face images
for which no such frontal face image is available, we simply employ face images with the same identity
and the smallest pose as their ground truth. Next, the standard stochastic gradient
descent (SGD) optimizer, with a weight decay of $5\times10^{-4}$ and a momentum of 0.9,
is utilized to train the entire network. The initial learning rate is set to 0.1.
The total number of training epochs is 30, and the epoch milestones are 5, 10, 15 and 20.
When the epoch reaches a milestone, the learning rate decays by a factor
of 0.5, 0.2, 0.1 and 0.1, respectively.

\begin{figure*}[htb]
	\setlength{\belowcaptionskip}{-0.4cm}
	\centering
	\includegraphics[width=1.0\textwidth]{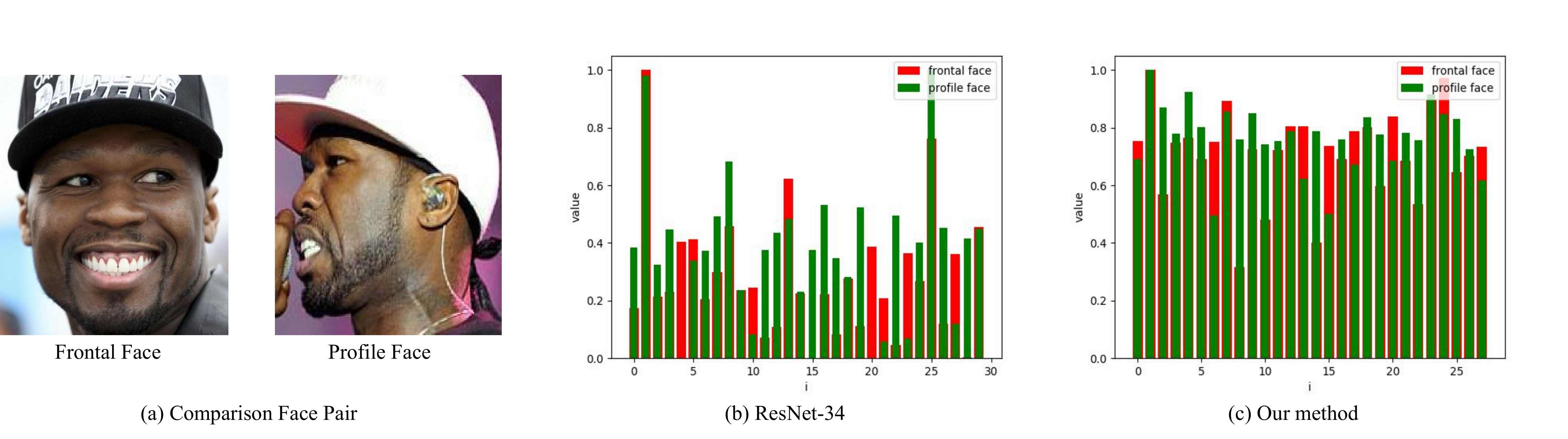}
	\caption{A visualization of the maximum elements in feature embeddings of a frontal face and a profile face.
		Red bars represent elements for the frontal face, while green bars represent the elements for the profile face. (a) Selected frontal 
		and profile face images. (b) Visualization of the embeddings extracted by the raw ResNet-34 model trained with only necessary classification loss.
		This indicates that feature embeddings between the frontal-profile face pairs not only share the common
		elements but also have their unique parts. (c) Visualization of our method, which significantly
		boosts the excavation of potential common elements across poses.
	}
	\setlength\belowcaptionskip{-1cm}
	\label{common_elements}
\end{figure*}

\begin{table*}
	\begin{center}
		\begin{tabular}{|p{2.0cm}<{\centering}|p{1cm}<{\centering}|p{1cm}<{\centering}|p{1cm}<{\centering}|
				p{2cm}<{\centering}|p{2cm}<{\centering}|}
			\hline
			Dataset & MSE&Prog.&APL &CFP & CPLFW\\
			\hline \hline
			\multirow{6}*{Ms-Subset} &-&-&- & $94.71\pm1.56$ & $83.93\pm1.96$ \\ 
			~ &\checkmark&-&- & $94.99\pm1.20$ & $82.89\pm2.41$ \\ 
			~ &-&\checkmark&- & $94.33\pm1.63$ & $83.99\pm2.20$ \\ 
			~ &-&-&\checkmark & $96.11\pm1.15$ & $84.41\pm2.52$ \\ 
			~ &\checkmark&\checkmark&- & $95.65\pm0.86$ & $83.63\pm2.31$ \\ 
			~ &-&\checkmark&\checkmark & \boldsymbol{$96.11\pm1.04$} & \boldsymbol{$85.30\pm2.05$} \\ 
			\hline
			\multirow{6}*{Ms-Refined} &-&-&- & $97.55\pm0.59$ & $87.47\pm1.61$ \\ 
			~ &\checkmark&-&- & $97.83\pm0.67$ & $86.93\pm1.65$ \\ 
			~ &-&\checkmark&- & $97.82\pm0.76$ & $88.64\pm1.37$ \\ 
			~ &-&-&\checkmark & $98.25\pm0.54$ & $88.69\pm1.83$ \\ 
			~ &\checkmark&\checkmark&- & $98.00\pm0.51$ & $87.42\pm1.71$ \\ 
			~ &-&\checkmark&\checkmark &\boldsymbol{$98.44\pm0.40$} & \boldsymbol{$89.37\pm1.87$} \\ 
			\hline
		\end{tabular}
	\end{center}
	\caption{Ablation study on the progressive module and APL. The frontal-profile face verification
		accuracy on the CFP dataset and face verification accuracy on the CPLFW dataset are reported here.}
	\label{AS0}
\end{table*}

\begin{table}
	\begin{center}
		\begin{tabular}{|p{3.0cm}<{\centering}|p{2cm}<{\centering}|p{2cm}<{\centering}|}
			\hline
			Method & CFP & CPLFW\\
			\hline
			Baseline & $94.99\pm1.20$ & $82.89\pm2.41$ \\
			\hline
			One block & $95.29\pm0.94$ & $83.26\pm2.09$ \\
			\hline
			Two block & $95.11\pm0.98$ & $83.55\pm2.40$ \\
			\hline			
			Three block & \boldsymbol{$95.65\pm0.86$} & \boldsymbol{$83.63\pm2.31$}\\
			\hline
		\end{tabular}
	\end{center}
	\caption{Ablation study on the number of blocks in the progressive module.
		Results include the frontal-profile face verification
		accuracy on the CFP dataset and face verification accuracy on the CPLFW dataset}
	\label{AS3}
\end{table}

\begin{table}
	\begin{center}
		\begin{tabular}{|p{3.0cm}<{\centering}|p{2cm}<{\centering}|p{2cm}<{\centering}|}
			\hline
			Gate setting & Accuracy & EER \\
			\hline
			Baseline & $94.99\pm1.20$ & $4.83\pm1.15$ \\
			\hline
			fixed $\gamma{\left(\theta_{i},\theta\right)}=1.0$ & $95.34\pm1.08$ & $4.27\pm1.09$ \\		
			\hline			
			unfixed & \boldsymbol{$95.65\pm0.86$} & \boldsymbol{$3.91\pm0.93$} \\
			\hline
		\end{tabular}
	\end{center}
	\caption{Ablation study on soft-gates. Accuracy and EER metrics of face verification on the CFP dataset are reported.}
	\label{AS4}
\end{table}

\begin{table*}
	\begin{center}
		\scriptsize
		\begin{tabular}{|p{4.0cm}<{\centering}|p{2cm}<{\centering}|p{2.5cm}<{\centering}|p{1.3cm}<{\centering}|}
			\hline
			Method & Training Data & Backbone & CFP-FP(\%) \\
			\hline
			Sengupta et al. \cite{author01} & CA(0.49M) & DCNN(1.7M) & 84.91 \\
			PIM \cite{author07} & MS(10M) & LightCNN(8.4M)  & 93.10 \\
			DR-GAN \cite{author53} & CA+MP(0.5M) & DCNN(13.0M) & 93.41 \\
			PFE \cite{author54} & MS(4.4M) & ResNet64(19.3M) & 93.34 \\
			Peng et al. \cite{author17} & CA+MP(0.5M) & CASIA-Net(1.7M) & 93.76 \\
			UV-GAN \cite{author46} & MS(4M) & ResNet27(35.1M) & 94.05 \\
			Yin et al. \cite{author20} & CA(0.49M) & p-CNN(1.8M) & 94.39 \\
			ArcFace \cite{author24} & MS(5.8M) & ResNet50(160.0M) & 95.56 \\
			Wang et al. \cite{author66} & MS(3.28M) & AttentionNet(89.3M) & 95.7 \\
			CircleLoss \cite{author55} & MS(3.6M) & ResNet34(21.3M) & 96.02 \\
			CurricularFace \cite{author56} & MS(5.8M) & ResNet100(248.0M) & 98.37 \\
			\hline
			\textbf{Our Method with MS subset} & & & \\
			ResNet-34 & MS(2.2M) & ResNet34(21.3M) & 94.71 \\
			Our method & MS(2.2M) & ResNet34(22.9M) & 96.11 \\
			\hline
			\textbf{Our Method with Refined MS} & & & \\
			ResNet-34 & MS(5.2M) & ResNet34(21.3M) & 97.55 \\
			Our method & MS(5.2M) & ResNet34(22.9M) & \textbf{98.4} \\
			\hline
		\end{tabular}
	\end{center}
	\caption{Comparisons in terms of frontal-profile verification accuracy on the CFP dataset. For the training data, we denote
		Ms-Celeb-1M, CASIA-WebFace, Multi-PIE as MS, CA and MP, respectively.}
	\label{CFP_FP}
\end{table*}

\begin{table*}
	\begin{center}
		\scriptsize
		\begin{tabular}{|p{4.0cm}<{\centering}|p{2cm}<{\centering}|p{2.5cm}<{\centering}|p{1.3cm}<{\centering}|}
			\hline
			Method & Training Data & Backbone & Accuracy(\%) \\
			\hline
			ML \cite{author59} & MS(7.03M) & MobileFaceNet(1.1M) & 82.44 \\
			MN \cite{author60} & MS(7.03M) & MobileFaceNet(1.1M) & 83.49 \\
			DC \cite{author61} & MS(7.03M) & MobileFaceNet(1.1M) & 84.01 \\
			CT \cite{author62} & MS(7.03M) & MobileFaceNet(1.1M) & 84.83 \\
			Co-Mining \cite{author63} & MS(7.03M) & MobileFaceNet(1.1M) & 85.70 \\
			Zhong et al. \cite{author65} & CA(0.67M) & ResNet50(43.6M) & 88.22 \\
			ArcFace \cite{author24,author66} & MS(3.28M) & AttentionNet(89.3M) & 88.78 \\
			Wang et al. \cite{author66} & MS(3.28M) & AttentionNet(89.3M) & 89.69 \\			
			CurricularFace \cite{author56} & MS(5.8M) & ResNet100(248.0M) & \textbf{93.13} \\
			\hline
			\textbf{Our Method with MS subset} & & & \\
			ResNet-34 & MS(2.2M) & ResNet34(21.3M) & 83.93 \\
			Our method & MS(2.2M) & ResNet34(22.9M) & 85.30 \\
			\hline
			\textbf{Our Method with Refined MS} & & & \\
			ResNet-34 & MS(5.2M) & ResNet34(21.3M) & 87.47 \\
			Our method & MS(5.2M) & ResNet34(22.9M) & \textbf{89.37} \\
			\hline
		\end{tabular}
	\end{center}
	\caption{Verification accuracy comparison on the CPLFW dataset. For the training data, we denote
		Ms-Celeb-1M, CASIA-WebFace, VGGFace2 respectively as MS, CA and VF2.}
	\label{CPLFW}
\end{table*}

\subsection{Ablation Study}
\subsubsection{Effectiveness of progressive blocks}
\par We first study the effectiveness of the progressive blocks.
The results are presented in Table \ref{AS0}.
It can be observed that the progressive module alone does not consistently boost the performance on both datasets.
This might be due to the lack of strong supervision such as MSE or APL, which would
help the network gain insight into how to drag the profile faces closer to the frontal faces in the
deep feature space. In fact, `MSE + progressive' and `APL + progressive'
outperform the `MSE-only' and `APL-only' settings, respectively. Through visualizing the deep feature space of the baseline 
ResNet-34 and ResNet-34 with the progressive blocks via t-SNE in Fig. \ref{Tsne}, we can observe that the progressive blocks 
help formulate more compact feature cluster in spite of pose variations.
These experimental results strongly verify the
effectiveness of our progressive module on face recognition across poses.

Next, we adopt 'MSE-only' as the baseline and conduct an experiment on the number of progressive blocks with different
pose splitting strategies and employ different amounts of progressive blocks. We divide the
face images into two groups (with the pose value within $\left[0^{\circ},45^{\circ}\right]$
and $\left[45^{\circ},90^{\circ}\right]$ ), three groups ($\left[0^{\circ},25^{\circ}\right]$,
$\left[25^{\circ},55^{\circ}\right]$, $\left[55^{\circ},90^{\circ}\right]$) and four groups ($\left[0^{\circ},20^{\circ}\right]$,
$\left[20^{\circ},40^{\circ}\right]$, $\left[40^{\circ},60^{\circ}\right]$, $\left[60^{\circ},90^{\circ}\right]$).
These splitting strategies respectively correspond to one block, two blocks and three blocks, which share the same basic
architecture. The results in Table \ref{AS3} demonstrate that it is more effective to segment the pose space and progressively
process face images of different poses.

Finally, to further verify the effectiveness of the progressive module, we remove the soft-gate methodology
and let $\gamma{\left(\theta_{i},\theta\right)}=1.0$, keeping the same parameters compared to our
progressive module. It can be observed from Table \ref{AS4} that our progressive module
achieves better performance. This indicates that the performance promotion in our approach is not
simply due to the increase of model complexity; instead, it is a result of the mechanism that
splits the pose space and processes each pose segment in a different manner.
\subsubsection{Effectiveness of APL}
To investigate the effectiveness of our proposed attentive pair-wise loss, we first conduct a qualitative analysis.
We randomly select a frontal face image and profile face image with the same identity from the CFP dataset, then visualize their
respective feature embeddings from the raw ResNet-34 model and our proposed method. We only consider the statistically top 20 maximum values
in the 512-dim feature vector, as these represent the most effective activation. The results in Fig. \ref{common_elements} reveal that
our proposed attentive pair-wise loss helps to boost the excavation of potential common activation across poses.

We then compare the performances of traditional MSE loss
and our proposed attentive pair-wise loss. The progressive module is removed to facilitate clean comparison. Following
Eq. \ref{total_loss}, MSE loss and APL are respectively combined with the cross entropy loss to train the backbone model with the
same settings. As is clear from the results in Table \ref{AS0}, our attentive pair-wise loss can significantly improve the
face recognition performance compared to the MSE loss. We further embed the progressive module into the entire network.
The results in Table \ref{AS0} show that the APL+Progressive setting yields the highest
performance, outperforming the MSE+Progressive setting.

\subsubsection{An Ablation Study on Balancing Parameters}
In this experiment, we study the effect of different parameters to balance the cross-entropy loss and
the pair-wise loss in Eq. \ref{total_loss}.
We carefully examine the possible parameters and assign the following weights to the pair-wise
losses: 0.01, 0.1, 0, 1, 2, and 10. We calculate the verification accuracy on both the CFP and CPLFW datasets and
present the results in Fig. \ref{mse_weight}. From the results, we can make the following observations:
First, the optimal balancing parameter for MSE loss is about 1.0, while that for our attentive
pair-wise loss is about 2.0. This can be explained by the fact that each element in the attention vector for APL is
actually smaller than 1.0 after maximum normalization; Therefore, the weight should be correspondingly somewhat larger.
Second, the best performance of our APL is significantly superior to all the settings of the MSE. This experiment
provides further evidence for the superior performance of APL.

\subsection{Comparisons with SOTA Methods}
\noindent\textbf{Results on CFP} Table \ref{CFP_FP}
presents the performance of recent methods on the CFP dataset, which covers almost all the categories of methods for face
recognition across poses discussed above. From the table we can observe that our proposed method significantly
boosts the face verification performance compared with the raw ResNet-34 model and surpasses that of the most recent methods
reported above. Notably, although our model is trained using a relatively smaller training set, it still outperforms
the CircleLoss \cite{author55}, which adopts the same backbone as our method but uses a much larger training set.
Furthermore, after we re-implement our method on a larger training set (the refined version of Ms-Celeb-1M), we
achieve higher performance than state-of-the-art methods, even though our backbone model is far more
light-weight.

\noindent\textbf{Results on CPLFW} Similar to the testing protocol on CFP, we follow the
official 10-fold cross validation protocol and test it on
the CPLFW dataset. We report the face verification accuracy and provide a comparison with recent methods in Table \ref{CPLFW}.
As can be seen from the table, our method achieves impressive improvement compared with the raw ResNet-34 model: specifically, 1.37$\%$
accuracy improvement on the subset of Ms-Celeb-1M and 1.9$\%\label{key}$ accuracy improvement on the refined
version of Ms-Celeb-1M. When compared with other methods, it can further be observed that our method is superior
to most of the above methods. If we were to implemented our method with a superior backbone and a larger-scale
training set, it would achieve better performance.

\section{Conclusion}
\label{sectionConclusion}
In this paper, we proposed a novel light-weight progressive module to model the nonlinear transformation from arbitrary faces to
frontal faces in the feature space. Our progressive module is not only effective in bridging the gap between the profile
faces and frontal faces, but can also flexibly cope with faces across various poses. Furthermore, we also
introduce an attentive pair-wise loss to guide the feature transformation progressing in the most effective direction.
Our method is light weight and easy to implement, increasing overall computational cost by only a small amount. Experiments
on the CFP and CPLFW datasets demonstrate that our proposed methods consistently brings significant
performance promotion for face recognition across poses.

\section{Acknowledgement}
\label{sectionAcknowledgement}
This work was supported by the National Natural Science Foundation of China under Grant 61702193, 
and the Fundamental Research Funds for the Central Universities of China under Grant 2019JQ01.

{\small
\bibliographystyle{ieee}
\bibliography{submit_bib}

\begin{thebibliography}{10}\itemsep=-1pt

\bibitem{author18}
K.~{Cao}, Y.~{Rong}, C.~{Li}, X.~{Tang}, and C.~C. {Loy}.
\newblock Pose-robust face recognition via deep residual equivariant mapping.
\newblock In {\em CVPR}, pages 5187--5196, 2018.

\bibitem{author37}
S.~{Chopra}, R.~{Hadsell}, and Y.~{LeCun}.
\newblock Learning a similarity metric discriminatively, with application to
  face verification.
\newblock In {\em CVPR}, volume~1, pages 539--546 vol. 1, 2005.

\bibitem{author46}
J.~{Deng}, S.~{Cheng}, N.~{Xue}, Y.~{Zhou}, and S.~{Zafeiriou}.
\newblock Uv-gan: Adversarial facial uv map completion for pose-invariant face
  recognition.
\newblock In {\em CVPR}, pages 7093--7102, 2018.

\bibitem{author24}
J.~{Deng}, J.~{Guo}, N.~{Xue}, and S.~{Zafeiriou}.
\newblock Arcface: Additive angular margin loss for deep face recognition.
\newblock In {\em CVPR}, pages 4685--4694, 2019.

\bibitem{author82}
C.~Ding and D.~Tao.
\newblock A comprehensive survey on pose-invariant face recognition.
\newblock {\em ACM TIST}, 7(3):1--42, 2016.

\bibitem{author80}
C.~Ding, K.~Wang, P.~Wang, and D.~Tao.
\newblock Multi-task learning with coarse priors for robust part-aware person
  re-identification.
\newblock {\em TPAMI}, pages 1--1, 2020.

\bibitem{author03}
Y.~Guo, L.~Zhang, Y.~Hu, X.~He, and J.~Gao.
\newblock Ms-celeb-1m: Challenge of recognizing one million celebrities in the
  real world.
\newblock {\em Electronic Imaging}, 2016:1--6, 2016.

\bibitem{author62}
B.~Han, Q.~Yao, X.~Yu, G.~Niu, M.~Xu, W.~Hu, I.~Tsang, and M.~Sugiyama.
\newblock Co-teaching: Robust training of deep neural networks with extremely
  noisy labels.
\newblock In {\em NIPS}, pages 8527--8537, 2018.

\bibitem{author73}
G.~{Huang}, M.~{Ramesh}, T.~{Berg}, and E.~{Learned-Miller}.
\newblock Labeled faces in the wild: A database for studying face recognition
  in unconstrained environments.
\newblock {\em Tech. rep.}, 10 2008.

\bibitem{author56}
Y.~{Huang}, Y.~{Wang}, Y.~{Tai}, X.~{Liu}, P.~{Shen}, S.~{Li}, J.~{Li}, and
  F.~{Huang}.
\newblock Curricularface: Adaptive curriculum learning loss for deep face
  recognition.
\newblock In {\em CVPR}, pages 5900--5909, 2020.

\bibitem{author60}
L.~Jiang, Z.~Zhou, T.~Leung, L.-J. Li, and L.~Fei-Fei.
\newblock Mentornet: Learning data-driven curriculum for very deep neural
  networks on corrupted labels.
\newblock In {\em ICML}, 2018.

\bibitem{author26}
M.~{Kan}, S.~{Shan}, H.~{Chang}, and X.~{Chen}.
\newblock Stacked progressive auto-encoders (spae) for face recognition across
  poses.
\newblock In {\em CVPR}, pages 1883--1890, 2014.

\bibitem{author32}
W.~{Liu}, Y.~{Wen}, Z.~{Yu}, M.~{Li}, B.~{Raj}, and L.~{Song}.
\newblock Sphereface: Deep hypersphere embedding for face recognition.
\newblock In {\em CVPR}, pages 6738--6746, 2017.

\bibitem{author72}
G.~Luo.
\newblock head-pose-estimation-and-face-landmark, 2016.

\bibitem{author61}
E.~Malach and S.~Shalev-Shwartz.
\newblock Decoupling "when to update" from "how to update".
\newblock In {\em NIPS}, pages 960--970, 2017.

\bibitem{author48}
I.~{Masi}, F.~{Chang}, J.~{Choi}, S.~{Harel}, J.~{Kim}, K.~{Kim}, J.~{Leksut},
  S.~{Rawls}, Y.~{Wu}, T.~{Hassner}, W.~{AbdAlmageed}, G.~{Medioni},
  L.~{Morency}, P.~{Natarajan}, and R.~{Nevatia}.
\newblock Learning pose-aware models for pose-invariant face recognition in the
  wild.
\newblock {\em TPAMI}, 41(2):379--393, 2019.

\bibitem{author17}
X.~{Peng}, X.~{Yu}, K.~{Sohn}, D.~N. {Metaxas}, and M.~{Chandraker}.
\newblock Reconstruction-based disentanglement for pose-invariant face
  recognition.
\newblock In {\em ICCV}, pages 1632--1641, 2017.

\bibitem{author10}
Y.~{Qian}, W.~{Deng}, and J.~{Hu}.
\newblock Unsupervised face normalization with extreme pose and expression in
  the wild.
\newblock In {\em CVPR}, pages 9843--9850, 2019.

\bibitem{author40}
F.~{Schroff}, D.~{Kalenichenko}, and J.~{Philbin}.
\newblock Facenet: A unified embedding for face recognition and clustering.
\newblock In {\em CVPR}, pages 815--823, 2015.

\bibitem{author01}
S.~{Sengupta}, J.~{Chen}, C.~{Castillo}, V.~M. {Patel}, R.~{Chellappa}, and
  D.~W. {Jacobs}.
\newblock Frontal to profile face verification in the wild.
\newblock In {\em WACV}, pages 1--9, 2016.

\bibitem{author54}
Y.~{Shi} and A.~{Jain}.
\newblock Probabilistic face embeddings.
\newblock In {\em ICCV}, pages 6901--6910, 2019.

\bibitem{author57}
Y.~{Shi}, X.~{Yu}, K.~{Sohn}, M.~{Chandraker}, and A.~K. {Jain}.
\newblock Towards universal representation learning for deep face recognition.
\newblock In {\em CVPR}, pages 6816--6825, 2020.

\bibitem{author55}
Y.~{Sun}, C.~{Cheng}, Y.~{Zhang}, C.~{Zhang}, L.~{Zheng}, Z.~{Wang}, and
  Y.~{Wei}.
\newblock Circle loss: A unified perspective of pair similarity optimization.
\newblock In {\em CVPR}, pages 6397--6406, 2020.

\bibitem{author53}
L.~{Tran}, X.~{Yin}, and X.~{Liu}.
\newblock Disentangled representation learning gan for pose-invariant face
  recognition.
\newblock In {\em CVPR}, pages 1283--1292, 2017.

\bibitem{author33}
H.~{Wang}, Y.~{Wang}, Z.~{Zhou}, X.~{Ji}, D.~{Gong}, J.~{Zhou}, Z.~{Li}, and
  W.~{Liu}.
\newblock Cosface: Large margin cosine loss for deep face recognition.
\newblock In {\em CVPR}, pages 5265--5274, 2018.

\bibitem{author81}
K.~Wang, C.~Ding, S.~J. Maybank, and D.~Tao.
\newblock Cdpm: Convolutional deformable part models for semantically aligned
  person re-identification.
\newblock {\em TIP}, 29:3416--3428, 2020.

\bibitem{author63}
X.~{Wang}, S.~{Wang}, H.~{Shi}, J.~{Wang}, and T.~{Mei}.
\newblock Co-mining: Deep face recognition with noisy labels.
\newblock In {\em ICCV}, pages 9357--9366, 2019.

\bibitem{author66}
X.~Wang, S.~Zhang, S.~Wang, T.~Fu, H.~Shi, and T.~Mei.
\newblock Mis-classified vector guided softmax loss for face recognition.
\newblock In {\em AAAI}, pages 12241--12248, 2020.

\bibitem{author20}
X.~{Yin} and X.~{Liu}.
\newblock Multi-task convolutional neural network for pose-invariant face
  recognition.
\newblock {\em TIP}, 27(2):964--975, 2018.

\bibitem{author71}
K.~{Zhang}, Z.~{Zhang}, Z.~{Li}, and Y.~{Qiao}.
\newblock Joint face detection and alignment using multitask cascaded
  convolutional networks.
\newblock {\em IEEE Signal Processing Letters}, 23(10):1499--1503, 2016.

\bibitem{author59}
Y.~{Zhang}, T.~{Xiang}, T.~M. {Hospedales}, and H.~{Lu}.
\newblock Deep mutual learning.
\newblock In {\em CVPR}, pages 4320--4328, 2018.

\bibitem{author07}
J.~{Zhao}, Y.~{Cheng}, Y.~{Xu}, L.~{Xiong}, J.~{Li}, F.~{Zhao}, K.~{Jayashree},
  S.~{Pranata}, S.~{Shen}, J.~{Xing}, S.~{Yan}, and J.~{Feng}.
\newblock Towards pose invariant face recognition in the wild.
\newblock In {\em CVPR}, pages 2207--2216, 2018.

\bibitem{author47}
J.~Zhao, L.~Xiong, P.~Karlekar~Jayashree, J.~Li, F.~Zhao, Z.~Wang,
  P.~Sugiri~Pranata, P.~Shengmei~Shen, S.~Yan, and J.~Feng.
\newblock Dual-agent gans for photorealistic and identity preserving profile
  face synthesis.
\newblock In {\em NIPS}, pages 66--76, 2017.

\bibitem{author29}
T.~Zheng and W.~Deng.
\newblock Cross-pose lfw: A database for studying cross-pose face recognition
  in unconstrained environments.
\newblock Technical Report 18-01, Beijing University of Posts and
  Telecommunications, February 2018.

\bibitem{author65}
Y.~{Zhong}, W.~{Deng}, M.~{Wang}, J.~{Hu}, J.~{Peng}, X.~{Tao}, and Y.~{Huang}.
\newblock Unequal-training for deep face recognition with long-tailed noisy
  data.
\newblock In {\em CVPR}, pages 7804--7813, 2019.

\end{thebibliography}
}

\end{document}